\documentclass[sigconf]{acmart}
\usepackage{microtype}
\usepackage{graphicx}
\usepackage{booktabs}
\usepackage{colortbl}
\usepackage{multicol}
\usepackage[most]{tcolorbox}
\usepackage{multirow}
\usepackage{float}
\usepackage{footmisc}
\usepackage{url}
\usepackage{makecell}
\usepackage[ruled,vlined]{algorithm2e}
\usepackage{algpseudocode}
\usepackage{subfigure}
\usepackage{mathtools}
\usepackage{nicematrix}
\usepackage{booktabs}
\usepackage{caption}

\usepackage{enumitem}
\usepackage{color}
\definecolor{Red}{RGB}{255,0,0}
\definecolor{Orange}{RGB}{237,125,49}
\definecolor{Green}{RGB}{0,128,0}
\definecolor{Blue}{RGB}{0,112,192}

\definecolor{'wit'}{HTML}{FBFBFB}
\definecolor{'gry'}{HTML}{EEEEEE}

\definecolor{'deep1'}{HTML}{C5E6F8} 
\definecolor{'shallow1'}{HTML}{E4F3FC} 
\definecolor{'deep2'}{HTML}{E5F5B7} 
\definecolor{'shallow2'}{HTML}{F3FADF} 

\definecolor{'deep3'}{HTML}{FFE5C6} 
\definecolor{'shallow3'}{HTML}{FFF2E3} 
\definecolor{'deep4'}{HTML}{FFD3CF} 
\definecolor{'shallow4'}{HTML}{FFEAE8}
\definecolor{'deep5'}{HTML}{D2D0F3} 
\definecolor{'shallow5'}{HTML}{E8E7F9} 

\algrenewcommand\algorithmicrequire{\textbf{Input:}}
\AtBeginDocument{%
  \providecommand\BibTeX{{%
    \normalfont B\kern-0.5em{\scshape i\kern-0.25em b}\kern-0.8em\TeX}}}

\setcopyright{acmcopyright}
\copyrightyear{2018}
\acmYear{2018}
\acmDOI{XXXXXXX.XXXXXXX}

\acmConference[Conference acronym 'XX]{Make sure to enter the correct
  conference title from your rights confirmation emai}{June 03--05,
  2018}{Woodstock, NY}
\acmPrice{15.00}
\acmISBN{978-1-4503-XXXX-X/18/06}

\begin{document}


\title[MAIC-UI: Making Interactive Courseware with Generative UI]{MAIC-UI: Making Interactive Courseware with Generative UI}

\author{Shangqing Tu$^{1}$, Yanjia Li$^{1}$, Keyu Chen$^{2}$, Sichen Zhang$^{3}$, Jifan Yu$^{1}$, Daniel Zhang-Li$^{1}$, Lei Hou$^{1}$, Juanzi Li$^{1}$, Yu Zhang$^{1}$, Huiqin Liu$^{1}$}
\affiliation{%
  \institution{$^{1}$Tsinghua University, $^{2}$Guangzhou University, $^{3}$Zhejiang University}
  \city{Beijing, Guangzhou, Hangzhou}
  \country{China}
\url{https://github.com/THU-MAIC/MAIC-UI}
}
\email{tsq25@mails.tsinghua.edu.cn}

\renewcommand{\shortauthors}{Tu, et al.}

\begin{abstract}
Creating interactive STEM courseware traditionally requires HTML/CSS/JavaScript expertise, leaving barriers for educators. While generative AI can produce HTML codes, existing tools generate static presentations rather than interactive simulations, struggle with long documents, and lack pedagogical accuracy mechanisms. Furthermore, full regeneration for modifications requires 200--600 seconds, disrupting creative flow. We present MAIC-UI, a zero-code authoring system that enables educators to create and rapidly edit interactive courseware from textbooks, PPTs, and PDFs. MAIC-UI employs: (1) structured knowledge analysis with multi-modal understanding to ensure pedagogical rigor; (2) a two-stage generate-verify-optimize pipeline separating content alignment from visual refinement; and (3) Click-to-Locate editing with Unified Diff-based incremental generation achieving sub-10-second iteration cycles. A controlled lab study with 40 participants shows MAIC-UI reduces editing iterations (4.9 vs. 7.0) and significantly improves learnability and controllability compared to direct Text-to-HTML generation. A three-month classroom deployment with 53 high school students demonstrates that MAIC-UI fosters learning agency and reduces outcome disparities---the pilot class achieved 9.21-point gains in STEM subjects compared to -2.32 points in control classes. Our code is available at \url{https://github.com/THU-MAIC/MAIC-UI}.
\end{abstract}



\begin{teaserfigure}
    \includegraphics[width=\textwidth]{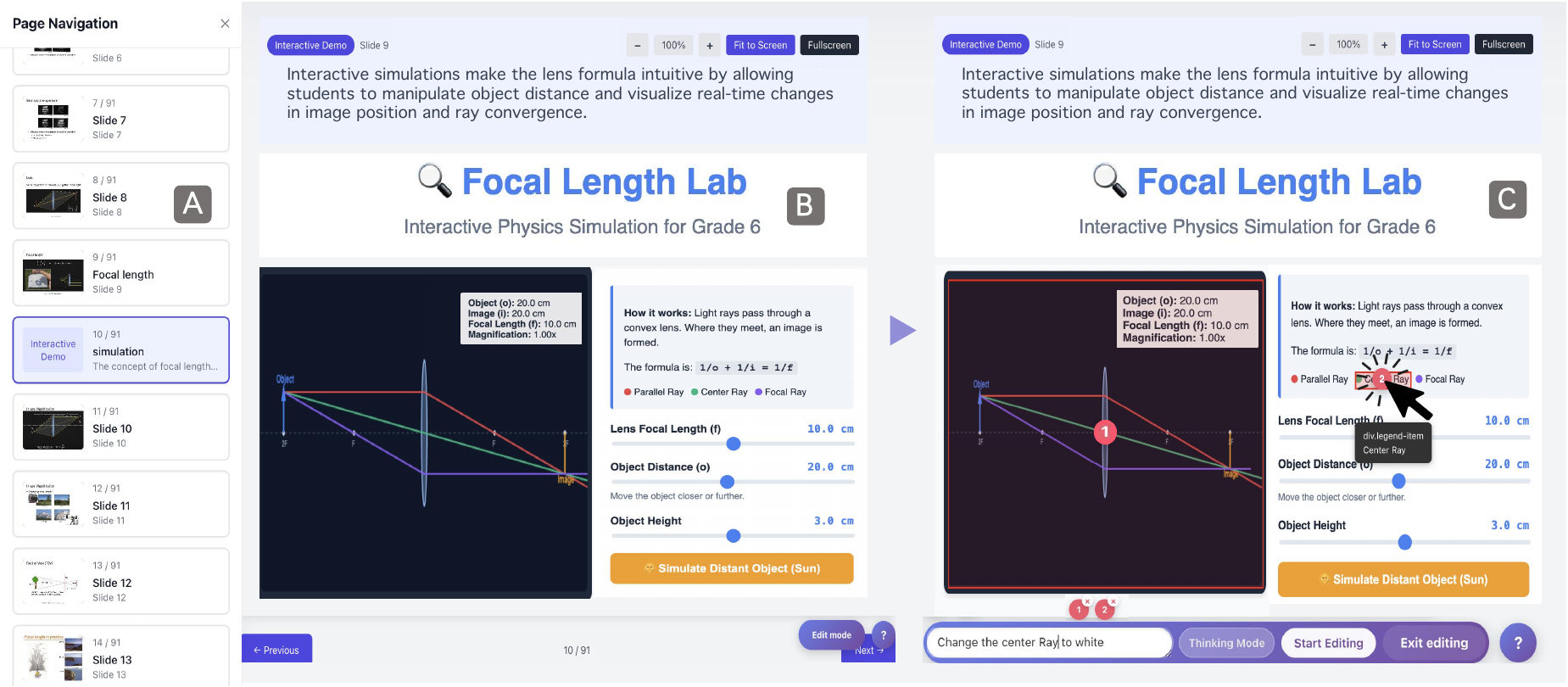}
    \caption{MAIC-UI enables zero-code creation and rapid editing of interactive courseware. In this example, a physics teacher creates an interactive simulation about lens focal length from a textbook chapter. (A) The Page Navigation panel displays the uploaded document pages, allowing teachers to browse content and select specific sections for generation. (B) The system generates an interactive simulation with a left panel showing the step-by-step procedure and a right panel with interactive controls for students to manipulate objects. (C) The Click-to-Locate Editing interface enables teachers to click on any element to select it, then describe desired changes in natural language, and apply Unified Diff-based updates in under 10 seconds.}
    \label{fig:system-pipeline}
\end{teaserfigure}

\maketitle


\section{Introduction}

Interactive courseware has emerged as a powerful tool for fostering student engagement and deepening conceptual understanding in STEM education~\cite{hoon2010effect,dalle2017interactive,su132413909,bell1987visual}. By enabling learners to manipulate parameters, observe real-time feedback, and explore procedural knowledge through hands-on experimentation, interactive simulations transform passive reception into active construction of knowledge~\cite{richards1992computer,fan2013enhancing,ben2022science}. However, creating such materials traditionally requires substantial programming expertise---teachers must write HTML, CSS, and JavaScript code to implement interactive components, debug cross-browser compatibility issues, and ensure visual polish~\cite{savoldelli2005barriers,gharib2023potential}. This technical barrier leaves many educators unable to realize their pedagogical visions, forcing them to rely on static presentations or pre-made simulations that may not align with their specific instructional goals.

Recent advances in generative AI have transformed content creation across domains, offering new possibilities for educational material development. Large language models can now generate code from natural language descriptions, potentially democratizing the creation of interactive web content~\cite{zeng2026glm,team2026kimi}. However, existing tools face significant limitations when applied to educational contexts: they often produce static text and images rather than interactive simulations~\cite{tu2023littlemu,wang2025educraft}, struggle to process long-form teaching materials such as textbooks and lecture slides~\cite{liu2024lost,bai2025longbench}, and lack mechanisms to ensure pedagogical accuracy---a critical requirement for educational content~\cite{yu2024kola,zhang2025siren}. Furthermore, the iteration cycles in current systems are painfully slow: when teachers request modifications, systems typically regenerate entire files, requiring 200--600 seconds per edit and disrupting creative flow~\cite{fakhoury2024llm,huang2024new}.

To understand these challenges from educators' perspectives, we conducted a formative study with six participants who had teaching experience. Our findings revealed four key challenges: \textbf{(F1) Knowledge accuracy concerns}---teachers worried that AI-generated content might misrepresent concepts or hallucinate wrong information; \textbf{(F2) Editing limitations}---participants expressed frustration with ambiguous natural language interfaces that made precise modifications difficult; \textbf{(F3) Passive learning experiences}---existing tools produced presentation-style content rather than interactive simulations that support learning-by-doing; and \textbf{(F4) Theory-practice gaps}---procedural knowledge was often presented as static descriptions rather than visualizable, explorable processes.

Based on these insights, we developed MAIC-UI (Making Interactive Courseware with Generative UI~\cite{leviathan2025generative}), a web-based authoring system that enables educators to create interactive courseware from textbooks, PPTs, and PDFs without programming expertise. MAIC-UI addresses the four challenges through four key design goals: \textbf{(DG1) Pedagogical scientificity and visual professionalism}---ensuring generated content is both factually accurate and aesthetically polished; \textbf{(DG2) Zero-code precise editing}---enabling teachers to make fine-grained adjustments without writing code; \textbf{(DG3) Active learning and personalized exploration}---supporting interactive components that foster student agency; and \textbf{(DG4) Procedural knowledge visualization}---making abstract processes concrete and explorable.

MAIC-UI employs three core technical innovations to achieve these goals. First, a \textit{structured knowledge analysis framework} with multi-modal understanding extracts pedagogical content from documents (up to 50 pages), ensuring accurate content understanding before generation. Second, a \textit{two-stage generate-verify-optimize pipeline} balances pedagogical accuracy with visual quality: Stage 1 creates content-aligned interactive simulations, while Stage 2 applies visual polish through layout verification, theme application, and animation smoothing. Third, \textit{Click-to-Locate natural language editing with Unified Diff-based incremental generation} enables teachers to select UI elements by clicking and describe changes in natural language, with the system applying only the necessary code patches in under 10 seconds---a 90\% reduction compared to full regeneration.

We evaluated MAIC-UI through a controlled user study with 40 participants and a three-month classroom deployment with 53 high school students. Results show that MAIC-UI reduces authoring time from days to minutes, achieves significantly faster edit response times compared to full regeneration baselines, and improves student engagement and learning outcomes. Teachers reported feeling empowered to create interactive content without programming, while students benefited from personalized, self-paced exploration experiences that traditional materials cannot provide.

In summary, this paper makes the following contributions:
\begin{itemize}
\item A formative study identifying key challenges educators face when creating AI-generated interactive courseware, informing four design goals for pedagogical content creation tools.
\item The design and implementation of MAIC-UI, a zero-code authoring system featuring structured knowledge analysis, two-stage generation with validation, and Click-to-Locate editing with Unified Diff-based incremental updates.
\item Empirical evidence from a controlled study (N=40) and classroom deployment demonstrating that MAIC-UI significantly reduces authoring burden, enables rapid iteration, and improves learning outcomes.
\end{itemize}

\section{Related Work}

\subsection{LLMs in Education}

Large language models are reshaping the production of educational content by simulating authentic dialogues and narrative logic to enhance learner immersion. For instance, Oak Story leverages LLM-mediated interactive narratives to construct immersive ecological learning environments~\cite{cheng2025Oak}, while TutorCraftEase supports automated generation of pedagogical questions and reduces teachers' burden~\cite{kangTutorCraftEaseEnhancingPedagogical2025}. Intelligent tutoring systems are also moving toward dynamic generation of personalized learning paths: LearnMate tailors learning plans for students with diverse backgrounds~\cite{wang2025LearnMate}, and GuideAI further incorporates physiological feedback to adapt instructional pacing~\cite{shukla2026GuideAI}. In project-based learning, LLM-assisted systems are transitioning from static guidance to dynamic interaction. While AutoPBL supports autonomous exploration through intelligent checkpoints~\cite{zhu2025AutoPBL}, another toolkit-based practice~\cite{li2025Unseen} improves students' engagement through multimodal creation.

However, despite improved content-production efficiency, existing educational LLM tools still struggle to ensure scientific rigor and visual professionalism, and often exhibit knowledge hallucination or misalignment between generated content and learning objectives~\cite{wang2025Large,park2024Empowering}. In addition, many systems still produce primarily presentation-oriented courseware and lack effective parametric interaction mechanisms that can genuinely support personalized learning-by-doing~\cite{zheng2024Charting,yoo2020learning}.

To address these issues, MAIC-UI establishes a two-stage workflow to improve pedagogical reliability and visual quality, and supports interactive component generation with real-time feedback to foster exploration through learner-controlled manipulation.

\subsection{AI Co-creation Authoring Tools}

In human-AI co-creation, a central design question is how to calibrate system agency so that AI can provide inspiration while preserving human decision authority. The COFI framework systematically characterizes this interaction design space~\cite{rezwana2023Designing}, and Reframer demonstrates how systems can balance emergent inspiration with user control~\cite{lawton2023Drawing}. Domain-specific authoring tools are also improving creative quality by integrating contextual knowledge. For example, AIdeation supports concept designers with efficient workflows for reorganizing reference materials~\cite{wang2025AIdeation}, while VRCopilot improves layout control in VR through intermediate representations such as wireframes~\cite{zhangVRCopilotAuthoring3D2024}. In UI-oriented creation, recent work has focused on automated evaluation and collaborative generation of high-fidelity components: DynaVis synthesizes interactive widgets from natural language~\cite{vaithilingam2024DynaVis}, and LLM-based evaluation can detect logical issues in UI mockups and provide optimization suggestions~\cite{duan2024Generating}.

Although current co-creation tools are effective at inspiration support, educators still face a single-change-affects-all dilemma in fine-grained revisions due to limited local editability. Slow feedback loops also interrupt creative flow and increase iteration cost~\cite{moruzzi2024Usercentered,lin2020It}. 

MAIC-UI addresses these gaps with Unified Diff-based incremental code updates, reducing file-modification latency from minutes to seconds, and with click-to-locate zero-code precise editing, enabling educators to revise specific interaction details directly while avoiding the instability of full regeneration.

\subsection{End-User Programming}

LLM-driven end-user programming (EUP) is evolving toward a responsible collaboration model that emphasizes guidance rather than one-shot code generation. CodeAid adopts cognitively supportive outputs such as pseudocode and explanatory comments~\cite{kazemitabaar2024CodeAid}, while SketchGPT introduces multimodal communication channels (e.g., sketches and language) for novice users~\cite{huang2025SketchGPT}. For understanding and modifying existing programs, interactive code-morphing approaches lower barriers via visual feedback: TweakIt allows non-experts to iteratively transform code behavior through real-time interaction~\cite{lauTweakItSupportingEnduser2021}, and Ply introduces clear boundary management for trigger-action programming~\cite{lim2018Ply}. EUP is also expanding into emerging domains such as mixed reality and robotics: agentAR supports rapid AR application construction through natural language~\cite{zhu2025agentAR}, and Alchemist simplifies robot behavior authoring into collaborative goal specification~\cite{karli2024Alchemist}. Recent work further explores support for developer's reflection in AI-assisted workflows~\cite{aveni2025Generative}.

Despite these advances, non-expert users still face barriers when modifying AI-generated artifacts. Existing EUP tools primarily support generating code from scratch but offer limited mechanisms for localizing and editing specific elements within generated programs~\cite{ma2025Scaffolding,yan2025Answering}. Users must either describe changes ambiguously in natural language or directly manipulate low-level code, creating cognitive barriers that violate the core EUP principle of letting users focus on \textit{what} to achieve rather than \textit{how} to implement it.

MAIC-UI addresses this gap through \textit{click-to-locate}: a mechanism that bridges visual elements and their underlying code by allowing users to select UI components through direct manipulation and describe desired changes in natural language. This approach enables teachers to iteratively refine interactive courseware without understanding DOM structures or CSS syntax, embodying the EUP vision of democratizing programming capabilities for domain experts.

\section{Formative Study}
\label{sc:formative}

To inform the design of MAIC-UI, we conducted a formative study to understand how educators experience AI-generated interactive courseware, the challenges they face during creation, and their perceptions of classroom integration. While prior research has explored AI-powered educational tools, limited understanding exists regarding teachers' experiences with generative UI systems that transform static teaching materials into interactive websites.

Our study was guided by the following research questions:

\begin{enumerate}
    \item How do educators navigate the interactive courseware creation process with AI assistance?
    \item What cognitive challenges do educators face when articulating instructional requirements to AI systems?
    \item How do educators perceive the integration of AI-generated interactive materials into classroom education?
\end{enumerate}

\subsection{Process}

We conducted semi-structured interviews with 6 participants recruited through local university networks. All participants were senior students from top Chinese universities with at least one teaching experience. Each participant completed an 1-hour hands-on session with MAIC-UI's initial version followed by a semi-structured interview exploring their creation experience, perceived learning costs, creative amplification, classroom integration concerns, and views on procedural knowledge visualization. We performed qualitative thematic analysis on the interview transcripts to identify recurring patterns and insights that informed our design goals. See Appendix~\ref{app:formative-methodology} for detailed recruitment criteria and participant quotes.

\subsection{Findings}

\subsubsection{F1: Knowledge Accuracy Concerns in AI-Generated Content}

Participants expressed concerns about knowledge accuracy and content fidelity in AI-generated courseware. P3 noted concerns about how the system ``represents knowledge,'' while P4 stated that ``sometimes the knowledge it produces is simply incorrect.'' Participants also identified mismatches between input content and generated output (P6: ``The generated website didn't include the content I had specified''). These findings reveal that teachers need mechanisms to ensure knowledge accuracy and content alignment. See Appendix~\ref{app:extended-quotes} for extended participant quotes.

\subsubsection{F2: Limitations of Current Editing Mechanisms}

Participants expressed the need for localized, granular editing without coding. P3 noted that ``modification and editing aren't that easy,'' while P5 estimated needing ``three to four'' iterations to achieve desired results. P6 described frustration with failed edits when the system ``didn't follow my instructions.'' These findings highlight a demand for ``what you see is what you get'' editing control that precisely targets specific elements without requiring code knowledge or multiple regeneration cycles.

\subsubsection{F3: Passive Courseware Fails to Engage Active Exploration}

Participants identified that traditional courseware promotes passive reception rather than active exploration. P4 described traditional PPT teaching as ``fixed content'' where ``students may find it boring.'' P2 emphasized that interactive tools ``offer students a buffet'' in an immediately actionable way, leading to higher participation. This reveals a need for hands-on, self-paced exploration opportunities.

\subsubsection{F4: Static Materials Cannot Bridge Theory-Practice Gaps}

Participants highlighted that students struggle to bridge the gap between conceptual knowledge and practical application. P4 explained that ``what they learn from textbooks and real-world scenarios have a gap,'' while P2 noted that ``problems are written elaborately, but the underlying knowledge is simple---some students cannot cross this chasm.'' These findings reveal a demand for dynamic visualizations that make abstract procedural steps concrete and visible.

\subsection{Design Goals}

Drawing on our formative study findings and learning theories, we identify four key design goals for an AI-powered interactive courseware authoring system:

\begin{itemize}
    \item \textbf{DG1: Ensure pedagogical scientificity and visual professionalism.}
    Address knowledge hallucinations and visual inconsistencies through a structured knowledge analysis framework and two-stage generate-verify-optimize pipeline, ensuring generated courseware maintains both scientific rigor and professional aesthetics.

    \item \textbf{DG2: Enable zero-code precise editing.}
    Support fine-grained localized adjustments through Click-to-Locate element selection and Unified Diff-based incremental generation, reducing edit response times from 200--600 seconds to under 10 seconds while preserving creative flow.

    \item \textbf{DG3: Foster active learning and personalized exploration.}
    Enable hands-on interaction with customizable parameters and real-time feedback, embodying Constructionist ``Learning by Doing'' principles to transform passive content consumption into active knowledge construction.

    \item \textbf{DG4: Visualize procedural knowledge to bridge learning gaps.}
    Generate dynamic visualizations at low cost to make abstract procedural logic concrete and visible~\cite{shu2025conversational}, reducing cognitive load and bridging the gap between conceptual understanding and practical application.
\end{itemize}

\section{The MAIC-UI System}

Drawing on findings from our formative study (Section~\ref{sc:formative}), we create a web-based authoring system, MAIC-UI, that enables educators to create interactive courseware from language instructions, PPTs, and PDFs without requiring programming expertise. Figure~\ref{fig:system-pipeline} illustrates the overall system architecture, which consists of three interconnected stages: Content Analysis, Two-Stage Generation, and Click-to-Locate Editing. Below, we present an illustrative scenario demonstrating how teachers use MAIC-UI in practice, followed by detailed descriptions of the core features and their implementation.

\subsection{Example Scenario}

Ms. Chen is a middle school physics teacher preparing a lesson on gravitational potential energy. She uploads a 45-page textbook chapter to MAIC-UI (Figure~\ref{fig:system-pipeline}, left). The system analyzes the content and identifies the subject area (Physics), key concepts (gravitational potential energy, mass, height), learning objectives, and the core formula $E_p = mgh$. Then the Two-Stage Generation Pipeline executes: Stage 1 creates an aligned simulation with a left panel showing the step-by-step process and a right panel with interactive controls; Stage 2 applies visual polish using a blue theme appropriate for physics content (Figure~\ref{fig:system-pipeline}, center). The complete generation takes 2.5 minutes. Noticing that the title appears plain, Ms. Chen clicks directly on it in the preview and types ``make this gradient red and bold.'' The system identifies the selected element, processes her instruction, and applies the change in 8 seconds (Figure~\ref{fig:system-pipeline}, right).

In the classroom, students interact with the generated simulation on their tablets. Each student can freely adjust the mass and height parameters, observing how the gravitational potential energy changes in real-time. One student comments: ``I can try different values without worrying about making mistakes. When I get stuck, the simulation shows me what happens---it's like having a tutor.'' This personalized exploration embodies the active learning principle (DG3), where students construct knowledge through hands-on experimentation rather than passive reception. The simulation also includes an animated visualization showing energy bars that grow and shrink as parameters change, making the abstract formula concrete and visible (DG4).


\begin{figure*}[t]
    \centering
    \includegraphics[width=\textwidth]{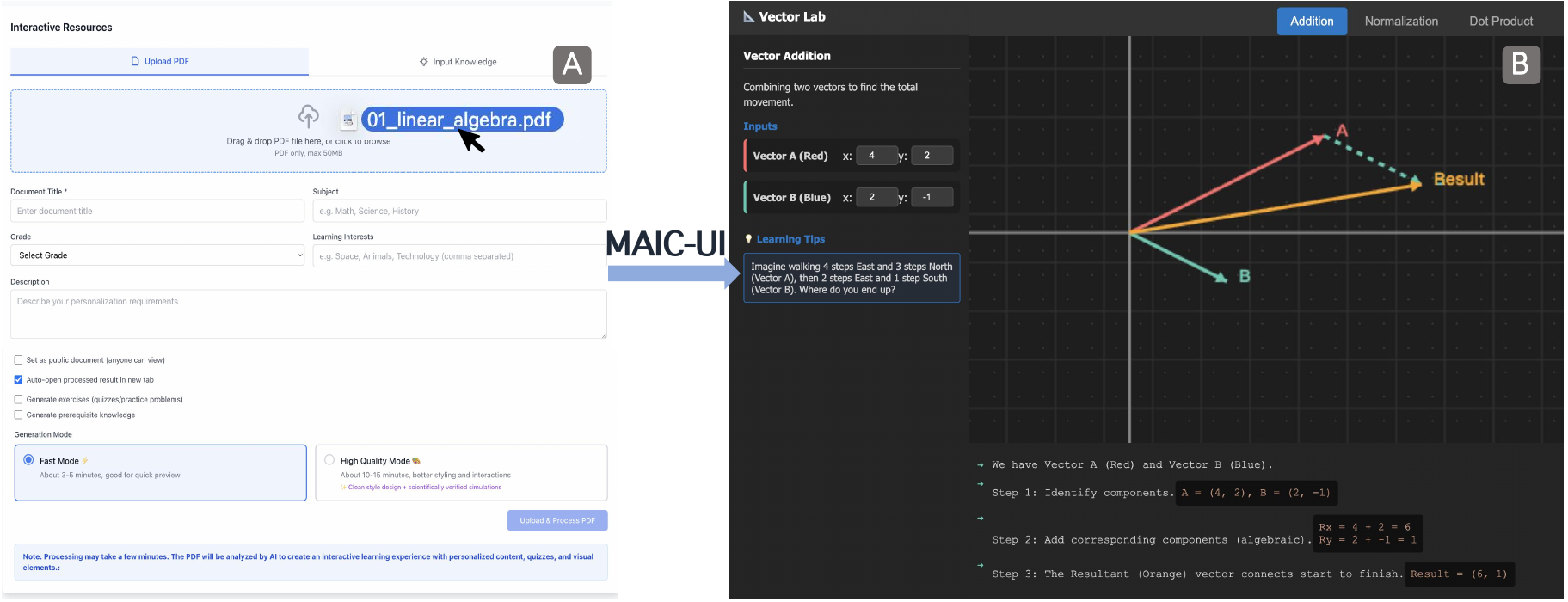}
    \caption{PDF Document Analysis and HTML Courseware Generation Process. (A) Teachers upload PDF documents containing pedagogical content. The system extracts page images and processes them through a vision-language model with structured analysis prompts to extract key concepts. (B) Based on the extracted structured knowledge, MAIC-UI applies subject-specific visual themes and generates interactive HTML courseware with pedagogically accurate content and professional styling.}
    \label{fig:pdf-generation}
\end{figure*}

\subsection{Key Features}

MAIC-UI integrates four key features that address the challenges identified in our formative study: structured knowledge analysis with multi-modal understanding (addressing F1, DG1), two-stage generation with validation (F1, DG1), and Click-to-Locate natural language editing with Unified Diff-based incremental generation (F2, DG2). These features collectively enable teachers to create pedagogically sound, visually professional interactive courseware while supporting student-centered active learning (F3, DG3) and procedural knowledge visualization (F4, DG4).

\subsubsection{Structured Knowledge Analysis with Multi-Modal Understanding}

MAIC-UI supports two input modes to accommodate different teacher workflows: (1) uploading PDFs for automatic concept extraction, or (2) directly entering structured pedagogical content. This dual-input design ensures that teachers can either leverage existing teaching materials or craft custom content from scratch.

\textbf{PDF-Based Input with Automatic Extraction.} Addressing participants' concerns about knowledge accuracy in AI-generated content (F1), MAIC-UI processes uploaded PDF documents through a multi-modal analysis pipeline (Figure~\ref{fig:pdf-generation}) that ensures accurate content understanding before generation. The system handles large documents (up to 50 pages) by extracting page images and sending them to a vision-language model with a structured analysis prompt. This approach prevents knowledge hallucinations and distinguishes MAIC-UI from simple Text-to-HTML systems that often produce pedagogically unsound content.

\begin{figure*}[t]
    \centering
    \includegraphics[width=\textwidth]{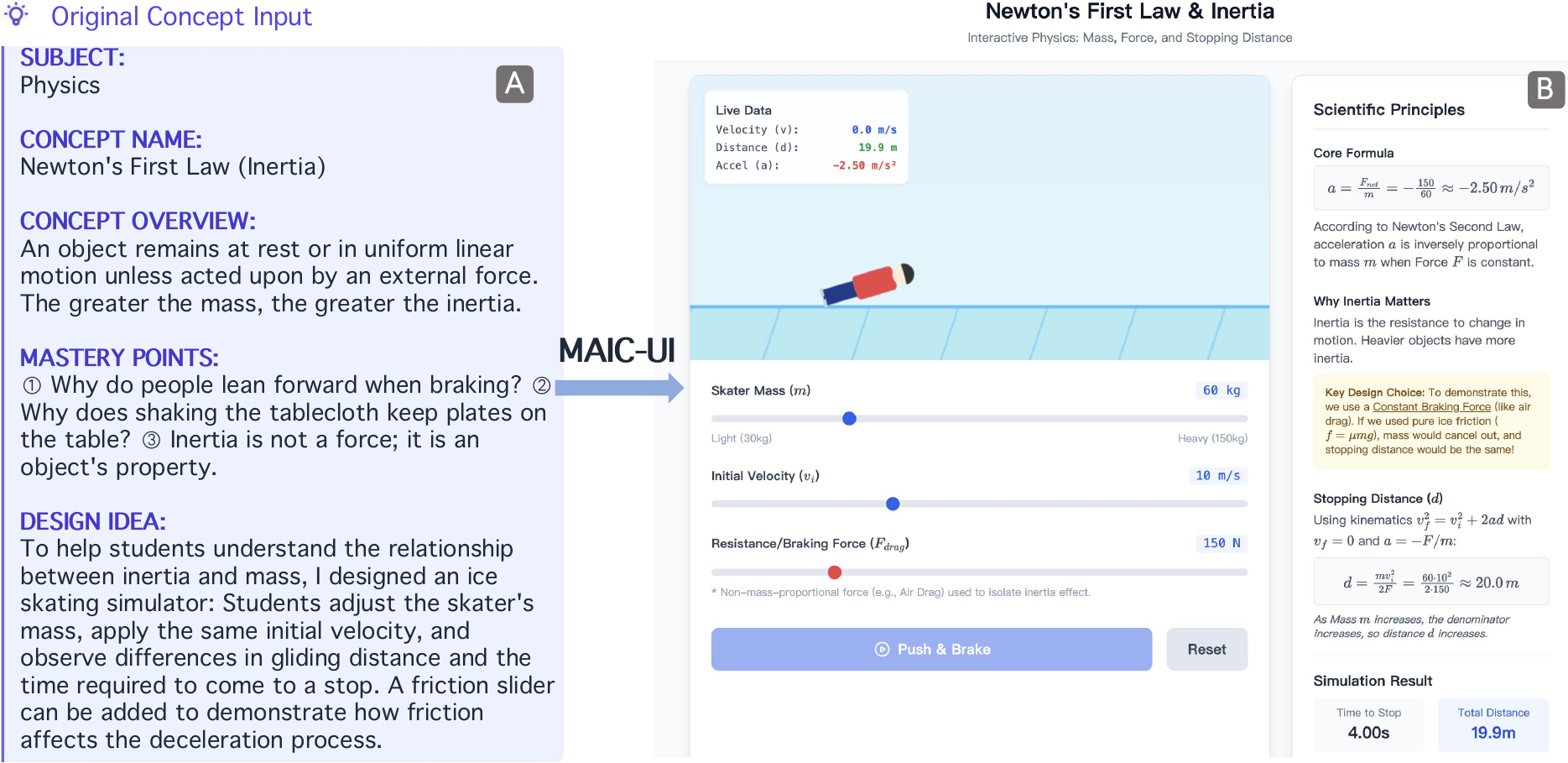}
    \caption{Concept-to-Interactive-Courseware Generation Pipeline. (A) Teachers input structured pedagogical content including subject, concept name, overview, mastery points, and design ideas. (B) MAIC-UI processes this input through the two-stage generation pipeline to produce interactive HTML courseware. The example shows a physics simulation for Newton's First Law where students can adjust skater mass, initial velocity, and friction parameters to observe how inertia affects stopping distance. The generated output includes interactive controls, real-time visualization, scientific principles sidebar, and simulation results.}
    \label{fig:concept-generation}
\end{figure*}

\textbf{Direct Concept Input.} Alternatively, teachers can directly input structured pedagogical content without uploading documents (Figure~\ref{fig:concept-generation}). This mode is particularly useful when teachers want to create courseware from scratch or when source materials are not available in PDF format. Teachers provide the subject, concept name, overview, mastery points, and design ideas through a structured form interface. This direct input bypasses the extraction step while still leveraging the same structured knowledge framework and generation pipeline.

Both input modes populate the same structured knowledge representation that drives subsequent generation. The analysis extracts six key fields that drive generation decisions: \textbf{Main Topics} identify the broad subject areas covered; \textbf{Key Concepts} capture specific terminology and principles students must master; \textbf{Learning Objectives} define measurable outcomes; \textbf{Prerequisite Knowledge} identifies foundational concepts; \textbf{Procedural Concepts} represent step-by-step processes that can be transformed into interactive simulations; and \textbf{Subject Area and Grade Level} enable theme selection. The procedural concepts field is particularly important---it captures processes that can become interactive simulations, distinguishing MAIC-UI from systems that only extract static knowledge.

Based on the extracted subject area, the system selects a visual theme from a predefined palette (e.g., blue tones for Physics, green for Biology, orange for Chemistry). Each theme defines primary and accent colors that ensure visual consistency throughout the generated courseware. This structured knowledge analysis framework ensures that all subsequent generation decisions are grounded in pedagogically accurate content understanding (DG1).

\subsubsection{Two-Stage Generation with Validation}

MAIC-UI employs a two-stage generation pipeline (Figure~\ref{fig:system-pipeline}, center panel) that balances pedagogical accuracy (F1, DG1) with visual professionalism (F1, DG1). This design addresses our formative finding that single-pass generation often produces either ``correct but ugly'' or ``beautiful but wrong'' outputs. As P4 emphasized, ``Sometimes the knowledge it produces is simply incorrect,'' highlighting the need for validation.

\textbf{Stage 1: Content-Aligned Interactive Simulations.} The first stage generates interactive simulations tightly aligned with extracted content. The layout consists of a left-side process panel displaying step-by-step procedural knowledge and a right-side simulation panel providing interactive visualization where students manipulate parameters and observe real-time effects. A coupling mechanism ensures that changes in the process panel automatically update the simulation state. The AI generates HTML/JavaScript code implementing this layout, and the output is validated to ensure interactive elements function correctly and respond to user input in real-time. If validation fails, the system attempts refinements, feeding validation errors back to the AI for regeneration.

\textbf{Stage 2: Layout Polish and Visual Refinement.} The second stage receives Stage 1 output and applies visual enhancements including layout verification, theme color application, typography improvements, and animation smoothing. An HTML validator checks for well-formed structure and proper syntax. Multiple refinement attempts are allowed to address any issues.

The pipeline implements graceful degradation: if Stage 1 fails, the system falls back to single-pass generation; if Stage 2 fails, basic styling is applied to Stage 1 output; if both fail, an emergency template is returned with a user-friendly message. This robust error handling ensures teachers always receive usable results.

\subsubsection{Click-to-Locate Natural Language Editing}

The Click-to-Locate Editing Module (Figure~\ref{fig:system-pipeline}, right panel) enables teachers to make fine-grained adjustments to generated content without writing code. This module directly addresses the editing limitations identified in our formative study (F2), where participants expressed frustration with existing tools. As P3 noted, ``Modification and editing aren't that easy...sometimes AI pretends to understand but gets it wrong.'' P5 estimated that achieving desired results required ``three to four times'' of iterations, while P6 described feeling stuck: ``I kept trying to change the website...but finally, it didn't follow my instructions.''

\textbf{Element Citation System.} The frontend implements a citation-based element selection system inspired by browser developer tools. When a teacher clicks on an element in the rendered preview, the system captures the element's structure and displays it in a side panel with an index number. Visual highlighting shows a border overlay on the selected element. This approach eliminates the need for teachers to understand CSS selectors or DOM traversal---they simply point and click. As P6 highlighted, the ability to directly connect visual elements to their underlying code ``reduces the uncertainty that comes from modifying through instructions alone,'' enabling true ``what you see is what you get'' editing.

\textbf{Natural Language to Code Translation.} Once an element is selected, teachers describe desired changes in natural language. The system constructs a prompt including the selected element, the modification instruction, and the full HTML source. The AI model processes this prompt and returns changes in a structured format for efficient application.

\textbf{Incremental Editing with Unified Diff.} A key innovation in MAIC-UI is using Unified Diff format for incremental code generation. This addresses the ``full regeneration bottleneck'' identified in our formative study, where systems output entire modified files resulting in wait times of 200-600 seconds. Instead, MAIC-UI outputs only changed lines plus minimal context, reducing output size by approximately 90\% and completing edits in under 10 seconds. The system implements fuzzy context matching to handle minor style drift, achieving high patch application rates even when the AI's expected context differs slightly from actual content.

The performance improvement enables teachers to stay in a creative flow state. As one participant reflected: ``Before, I would lose my train of thought waiting 5-10 minutes for each change. Now it's instant---I can iterate like I'm sketching on paper.'' This rapid iteration capability is essential for supporting teachers' pedagogical creativity without technical friction (DG2).

\subsection{Implementation}

MAIC-UI is implemented as a full-stack web application designed for multi-user interaction. The frontend is built with React 18 and TypeScript~\cite{bierman2014understanding}, providing interfaces for visual editing with split-pane preview, element citation display, and natural language chat commands. We chose React for its component-based architecture, which aligns well with our modular courseware generation approach. The backend uses Python FastAPI with SQLite~\cite{gaffney2022sqlite} for data storage, implementing service layers for content generation and editing operations. The system integrates Zhipu AI's GLM-4.7~\cite{zeng2025glm} for text generation and GLM-4.6V~\cite{hong2025glm} for multi-modal analysis, with fallback support for other endpoints~\cite{team2024gemini,achiam2023gpt}.

The Click-to-Locate editing module implements a DOM-aware element citation system. When a user clicks on a rendered element, the frontend captures the element's XPath~\cite{benedikt2009xpath} and CSS selector, then displays the corresponding HTML snippet in a side panel. The natural language instructions are sent to the backend along with the full HTML context and selected element. The AI model processes this prompt and returns changes in Unified Diff format~\cite{nugroho2020different}.


See Appendix~\ref{app:implementation} for complete API specifications, AI model configurations, prompt templates, and code-level implementation details.

\section{Evaluation}

\paragraph{Ethics.}

This evaluation received approval from the Institutional Review Board (IRB) of the authors' institution. All participants provided informed consent before joining the study, and demographic data (e.g., discipline background and prior teaching experience) were recorded anonymously.

\subsection{Research Questions}

To evaluate MAIC-UI in both controlled and authentic settings, we combine a lab user study on authoring performance with an in-class deployment on learning outcomes. The following research questions guide this evaluation.

\begin{itemize}
    \item \textbf{RQ1 (Lab User Study):} To what extent does MAIC-UI improve the pedagogical correctness and visual professionalism of generated instructional webpages compared with direct Text-to-HTML generation?
    \item \textbf{RQ2 (Lab User Study):} To what extent does MAIC-UI support fine-grained, predictable, and efficient editing for users without programming experience?
    \item \textbf{RQ3 (In-Class Deployment):} How can students' learning agency be fostered in traditional classroom settings?
    \item \textbf{RQ4 (In-Class Deployment):} How can gaps in learning outcomes among students with different learning abilities be mitigated?
\end{itemize}

\subsection{Lab User Study}
\label{sec:lab_user_study}

\subsubsection{Conditions}

We compare two conditions to isolate the contribution of MAIC-UI's full pipeline, especially its intermediate structuring and targeted editing support.

\begin{itemize}
    \item \textbf{Baseline} A simplified version of MAIC-UI in which both initial prompts and subsequent revision instructions were sent directly to the AI model for HTML generation. Unlike the full MAIC-UI pipeline, this baseline did not include any intermediate processing, structured decomposition, or validation before output.

    \item \textbf{MAIC-UI} The full version of the MAIC-UI including complete step of generation.
\end{itemize}

\subsubsection{Participants}

We recruited 40 graduate students with prior teaching practicum experience as proxy instructors for the controlled authoring evaluation. Their disciplinary backgrounds included Social Sciences and Management (\(n=2\)), Computer Science (\(n=7\)), Basic Sciences (\(n=8\)), and Engineering (\(n=23\)). Participants were evenly assigned to the two study conditions, with 20 in the experimental group and 20 in the control group.

\subsubsection{Authoring Tasks}

We designed the study task to reflect the complete MAIC-UI workflow, spanning both teacher preparation and student learning. Each participant engaged in both interactive courseware authoring and the review of webpages created by others, allowing them to evaluate the system from the perspectives of both content creators and learners.

To ensure realism and pedagogical validity, we used authentic STEM teaching materials collected from real educational settings ranging from primary school to graduate-level courses. These materials were collected with IRB approval and consent from both the instructors and the student authors. Each task package consisted of a 20--30 page slide deck and a corresponding teaching outline, covering subjects such as science, chemistry, biology, mathematics, and geography, and was assigned to participants based broadly on their disciplinary backgrounds.

\subsubsection{Procedure}

Each session lasted approximately 80 minutes and consisted of four stages: material familiarization, interactive courseware authoring, peer review, and a post-study interview.

\paragraph{\textbf{Stage 1: Material Familiarization.}}
Participants first spent approximately \textbf{10 minutes} familiarizing themselves with the assigned teaching materials. This stage helped them develop an initial understanding of the lesson content and how it might be organized into an interactive instructional webpage.

\paragraph{\textbf{Stage 2: Interactive Courseware Authoring.}}
Participants then spent approximately \textbf{45 minutes} using the assigned system to create and modify interactive instructional webpages. An initial version had been pre-generated from the slide materials for efficiency, but participants were still required to upload PDF or text materials themselves so that they could experience the full document-based generation workflow before further revising the webpage.

\paragraph{\textbf{Stage 3: Peer-work Review.}}
Participants then spent approximately \textbf{10 minutes} reviewing and interacting with webpages created by other participants from a learner's perspective.

\paragraph{\textbf{Stage 4: Post-task Questionnaire and Interview.}}
Participants then completed a post-task questionnaire followed by a semi-structured interview, in which we explored their perceptions of the system's usability, generation quality, and overall value for teaching. See Appendix~\ref{app:questionnaire} for the complete questionnaire and interview guide.

\begin{table}[t]
\centering
\caption{Comparison of editing behavior between MAIC-UI and the baseline condition (\(n=20\) per condition).}
\label{tab:study1_quantitative}
\begin{tabular}{lcccc}
\toprule
\textbf{Method} & \textbf{M} & \textbf{SD} & \textbf{Mdn} & \textbf{IQR} \\
\midrule
MAIC-UI  & 4.90 & 2.88 & 4.50 & 2.75--7.00 \\
Baseline & 7.00 & 2.20 & 7.00 & 5.00--9.00 \\
\midrule
\multicolumn{5}{p{0.95\linewidth}}{\footnotesize Mann--Whitney \(U\) test: \(U = 113.0\), \(p = 0.019\), effect size \(r = 0.37\).} \\
\bottomrule
\end{tabular}
\end{table}

\subsubsection{Results}

In this subsection, we report quantitative and qualitative findings from the lab study. We first present editing-behavior outcomes, and then summarize questionnaire and interview results to provide a more complete view of usability and perceived value.

\paragraph{\textbf{Editing Accuracy and Learnability}}

As shown in Table~\ref{tab:study1_quantitative}, editing behavior differed noticeably across the two conditions. Participants in the MAIC-UI condition generally completed refinement in fewer iterations, most often within about 3--7 rounds, whereas those in the baseline condition more commonly needed around 5--9 rounds. This suggests that MAIC-UI supported a more efficient and stable path from initial draft to target outcome (supporting RQ2). One explanation is that the Click-to-Locate interaction enabled localized, intention-aligned revisions, thereby reducing the need for repeated global regeneration (Figure~\ref{fig:system-pipeline}, right).

\begin{figure*}
    \centering
    \includegraphics[width=1\linewidth]{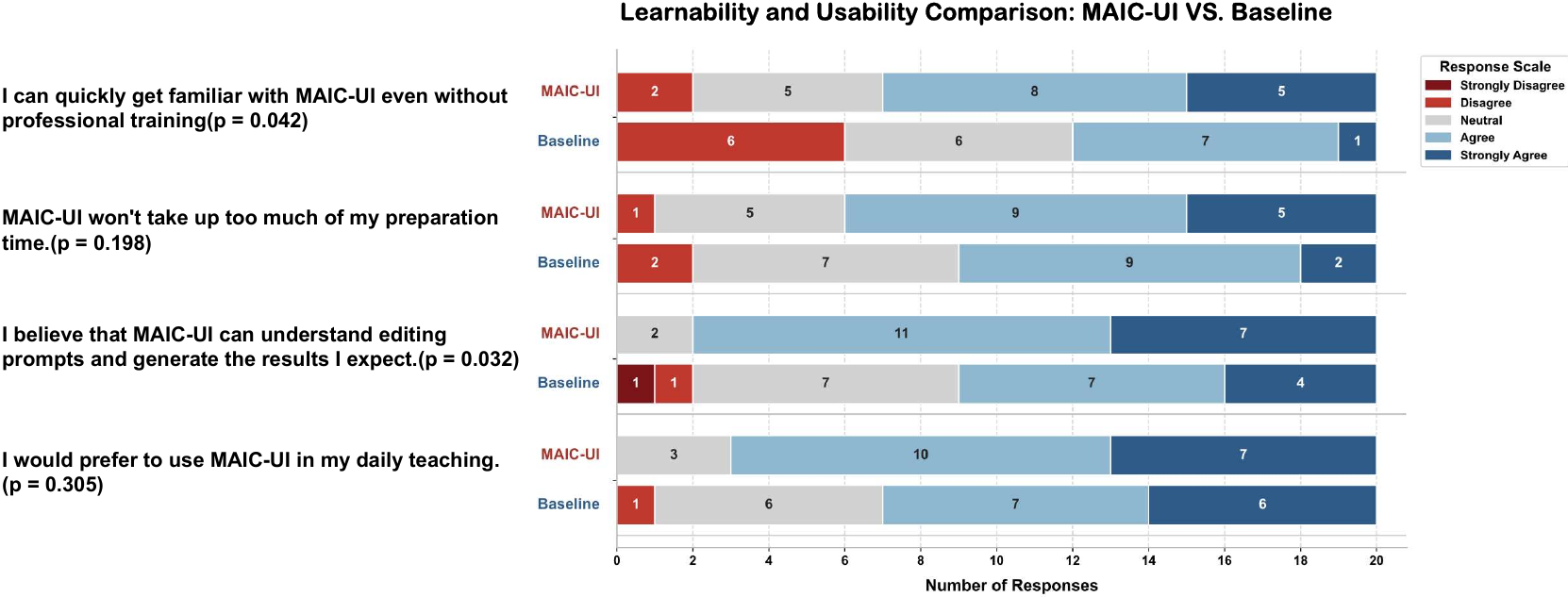}
    \caption{Questionnaire results comparing MAIC-UI and the baseline in the lab user study (\(n=20\) per condition). Stacked bars show response distributions for the four core items: learnability, time cost, editing controllability, and usage preference.}
    \label{fig:qs1}
\end{figure*}

\begin{figure*}
    \centering
    \includegraphics[width=0.95\linewidth]{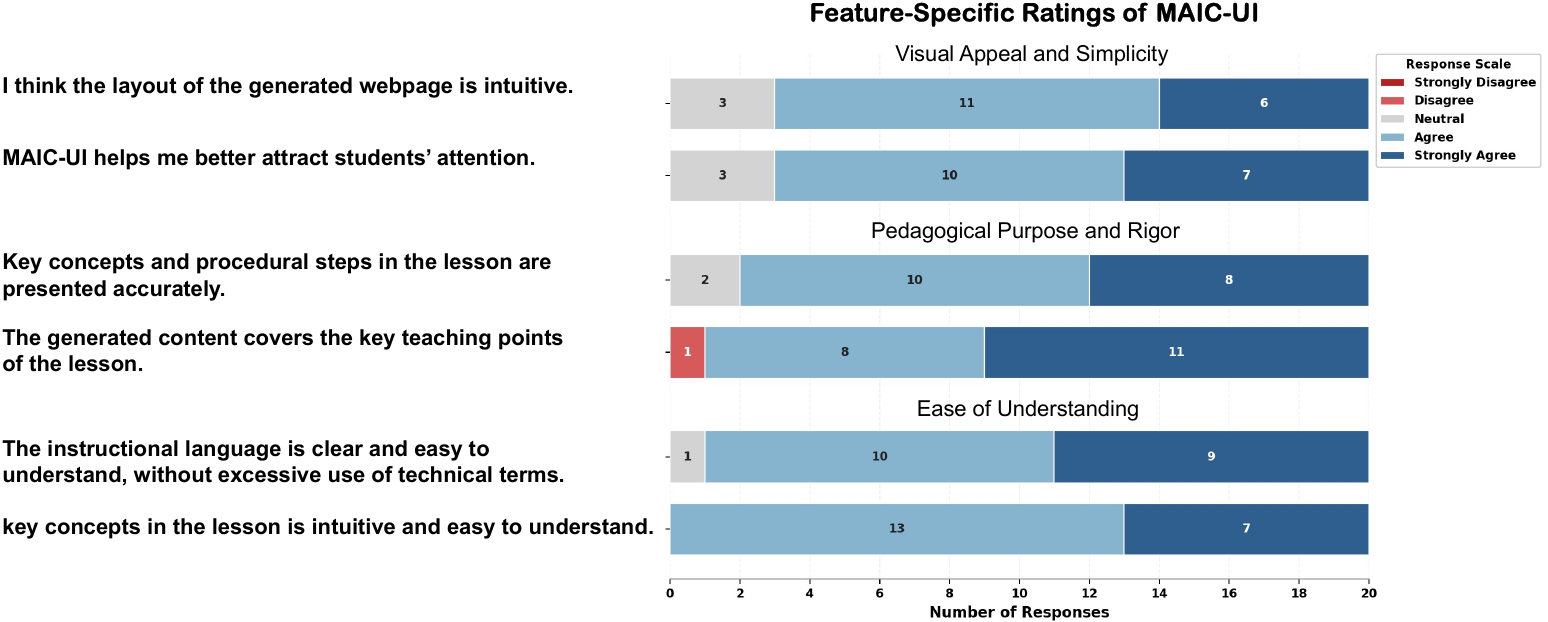}
    \caption{The six items summarize participants' judgments along three dimensions: visual appeal and simplicity, pedagogical purpose and rigor, and ease of understanding.}
    \label{fig:rq1_items}
\end{figure*}

\paragraph{\textbf{Questionnaires}}

For each learnability and usability item, we encoded five-point Likert responses as ordinal scores from 1 (strongly disagree) to 5 (strongly agree), and compared MAIC-UI and the baseline using two-sided Mann--Whitney \(U\) tests (\(n=20\) per condition). We used \(p<0.05\) as the threshold for statistical significance. As illustrated in Figure~\ref{fig:qs1}, the results show that MAIC-UI was rated significantly higher than the baseline on two of the four items, while the remaining two items did not reach statistical significance. Specifically, Item 1 (\(M=3.80\) vs. \(3.15\), \(p=0.042\)) and Item 3 (\(M=4.25\) vs. \(3.60\), \(p=0.033\)) showed significant differences, whereas Item 2 (\(M=3.90\) vs. \(3.55\), \(p=0.199\)) and Item 4 (\(M=4.20\) vs. \(3.90\), \(p=0.305\)) did not. For the two non-significant items, ratings in both conditions were already relatively high, suggesting a possible ceiling effect that limited further between-condition separation. Taken together, these results indicate that MAIC-UI better supports users without programming experience in performing editing in a more predictable and efficient manner, especially in helping them become familiar with the system more quickly and obtain results that better aligned with their editing intentions (supporting RQ2).

To further examine webpage quality beyond editing experience, we analyzed how participants in the MAIC-UI condition perceived the generated webpages themselves. Experimental-group participants rated all six RQ1-specific items positively, including layout intuitiveness (\(M=4.15\)), attention attraction (\(M=4.20\)), accurate presentation of key concepts and procedural steps (\(M=4.30\)), coverage of key teaching points (\(M=4.45\)), clarity of instructional language (\(M=4.40\)), and intuitive presentation of key concepts (\(M=4.35\)) (Figure~\ref{fig:rq1_items}). All six items received mean ratings above 4.0, indicating that MAIC-UI achieved consistently positive evaluations in terms of visual quality, pedagogical soundness, and accessibility (supporting RQ1).

\paragraph{\textbf{Interview Findings.}}

To complement the behavioral and questionnaire results, we analyzed participants' semi-structured interviews and identified two themes.

\textit{Theme 1. MAIC-UI helps translate ideas into workable instructional presentations.} Participants described being able to offload initial structuring work to the system. P1 noted that while making slides manually requires finding images and typesetting formulas, MAIC-UI generates ``the layout and even questions,'' feeling ``very efficient overall.'' P17 highlighted a case where the system ``generated a presentation using a running track to illustrate the relationship between linear velocity and angular velocity''---described as ``a very brilliant classroom introduction.'' P8 estimated efficiency improvements of ``around three times,'' noting that ``you only need to give it the goal you want, and it can directly make it for you.''

\textit{Theme 2. Effective editing requires explicit intentions but is learnable.} P1 emphasized that ``if you describe to the AI in more detail and with more precision, it can often give you the effect you want most.'' Participants viewed this skill as developable rather than expert-only---P4 noted that ``if you do it a few more times or practice more diligently, the results will get better,'' and P1 suggested that with documentation and training, even novice users could find the system ``quite easy to pick up.''

\subsection{In-Class Deployment}

\subsubsection{Classroom Context}

Beyond the controlled user study, we conducted a three-month in-the-wild deployment of MAIC-UI in a second-year class at a county-level public high school in China. Rather than aiming for a strictly controlled evaluation, this deployment sought to situate MAIC-UI within an authentic and sustained teaching context, allowing us to examine its practical role in everyday classroom use. Specifically, we explored whether MAIC-UI could foster students' learning agency in traditional classrooms and whether its long-term use could help reduce disparities in learning outcomes among students with different levels of academic ability.

The grade comprised 11 classes, from which one class with 53 students was selected as the pilot class based on the school's instructional arrangements and deployment feasibility. All students in this class were enrolled in the physics--chemistry--biology elective track, a science-oriented subject combination broadly aligned with STEM-related learning in the Chinese upper-secondary curriculum. Prior to the deployment, the classroom was equipped with the basic hardware infrastructure required for daily use of MAIC-UI, including tablets and charging cabinets.

\subsubsection{Study Procedure}

After the November monthly examination in 2025(pre-exam), Class C1 officially adopted MAIC-UI and continued using it until the final examination in February 2026(post-exam), resulting in a deployment period of approximately three months. Before each class, the teacher uploaded courseware created with MAIC-UI to the system. During class, in addition to listening to the teacher's instruction, students could also use tablets to independently interact with the embedded interactive components to support their understanding of the course content.

As this deployment took place in an authentic and ongoing classroom setting, we did not structure it as a formal study under strictly controlled conditions. Instead, we addressed the two research questions by comparing changes in students' performance across the two examinations and by conducting interviews with a subset of students and teachers after the deployment.

\begin{figure}[t]
    \centering
    \includegraphics[width=\linewidth]{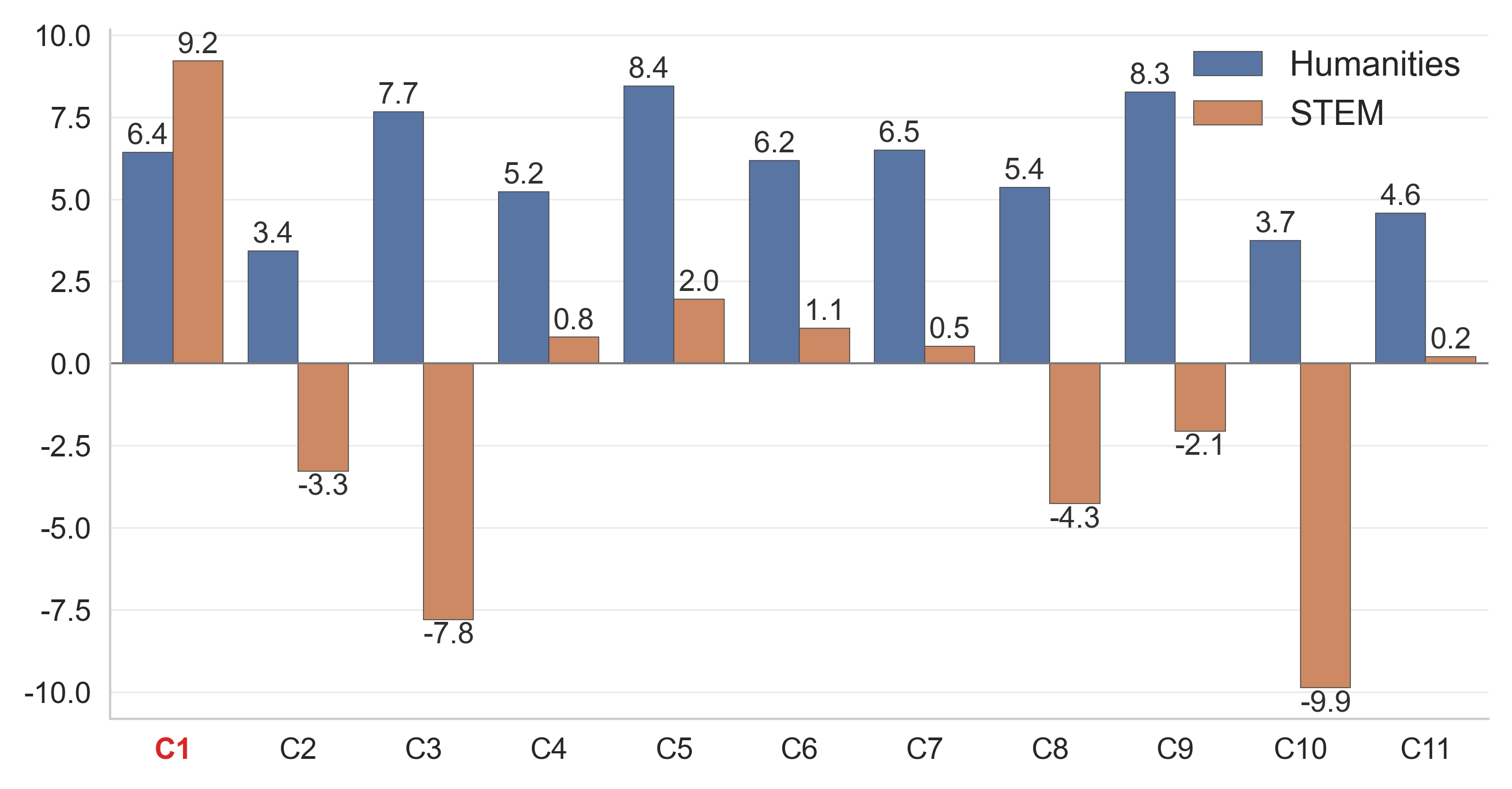}
    \caption{Score gains in STEM and humanities across classes from the November 2025 monthly examination to the February 2026 final examination.}
    \label{fig:score-gains-by-class}
\end{figure}

\begin{figure}[t]
    \centering
    \includegraphics[width=\linewidth]{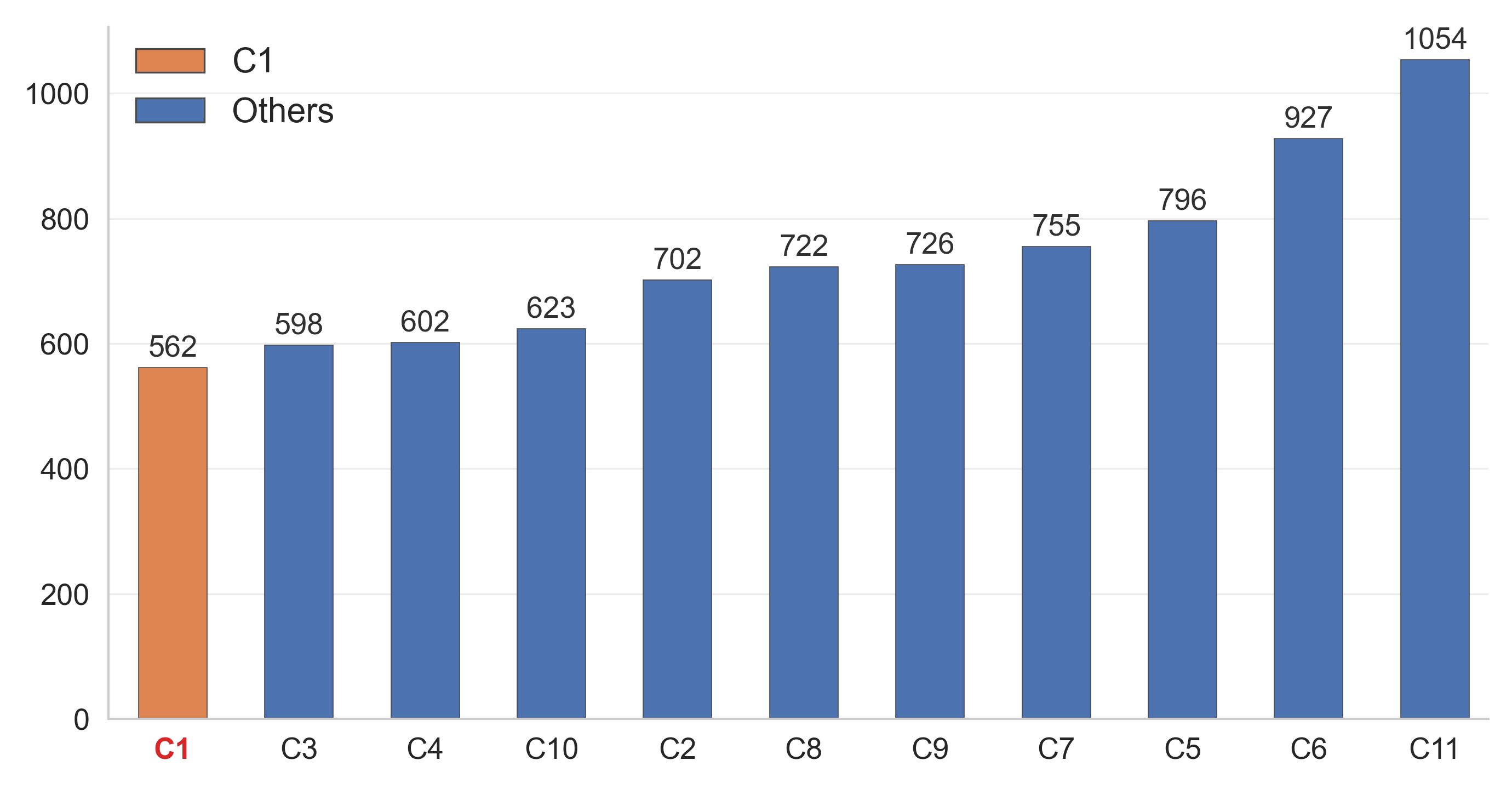}
    \caption{Variance of STEM score gains across classes over the same period. C1 denotes the MAIC-UI pilot class.}
    \label{fig:stem-variance-by-class}
\end{figure}

\subsubsection{Results}
In this section, we present quantitative outcomes from the classroom deployment. We report overall score gains, the distribution of gains within class, and performance changes among lower-performing students.

\paragraph{\textbf{Overall Score Gains}}

We first compared score gains between Class C1 and the other classes in the same grade from the pre-exam to the post-exam. Given that the classroom deployment of MAIC-UI primarily supported STEM subjects, including physics, chemistry, and biology, we separately calculated score gains for STEM subjects and humanities subjects.

The results showed that, in STEM subjects, Class C1 exhibited the most substantial improvement, with an average gain of 9.21 points, while most other classes showed only modest gains or declines. Exploratory student-level comparison indicated that C1 (\(n=53\), \(M=9.21\)) had higher STEM gains than the other classes (\(n=493\), \(M=-2.32\); Mann--Whitney \(U=16691.5\), \(p<0.001\), \(r=0.14\)).

In humanities subjects, Class C1 also showed a positive gain, with an average increase of 6.43 points, although this was not the highest in the grade. Compared with STEM subjects, the relative advantage of Class C1 in humanities was less pronounced. This difference is broadly consistent with the design focus of MAIC-UI, whose strengths are more likely to emerge in STEM learning content that benefits from dynamic presentation, interactive manipulation, and support for understanding processes.

\paragraph{\textbf{Variance of Score Gains}}

Beyond overall score gains, we examined the variance of score changes across classes as an indicator of how evenly improvement was distributed among students. As shown in Figure~\ref{fig:stem-variance-by-class}, Class C1 exhibited the lowest score-gain variance in the grade (\(562\)), compared with 598--1054 for the other classes.

This pattern suggests that learning gains in C1 were more evenly distributed across students, rather than being concentrated among a small subset of higher-performing learners. In relation to RQ4, this provides descriptive evidence that gaps in learning outcomes between students with different ability levels may have been mitigated. This pattern is also consistent with the mechanism proposed in our method: by aligning animated demonstrations with procedural knowledge in the generated interactive pages, the system externalizes content that would otherwise require students to mentally infer and internalize on their own, thereby making key problem-solving steps more directly learnable and potentially more accessible to lower-performing students.

\paragraph{\textbf{Gains Among Lower-Performing Students}}

We further examined students in the bottom 25\% of the November 2025 monthly examination. In STEM subjects, the bottom 25\% students in Class C1 (\(n=14\)) showed a larger average score gain than their counterparts in the other classes (\(n=137\), 15.46 vs.\ 12.42), a higher proportion of positive gains (78.6\% vs.\ 63.5\%), and a lower variance of score gains (251.02 vs.\ 671.34).

These results suggest that the benefits of MAIC-UI were not limited to students who were already more self-motivated or academically stronger. Instead, the observed pattern indicates that its support may extend to a broader range of learners, including those who initially performed at a lower level.

\subsubsection{Post-Deployment Interviews}

After the three-month classroom deployment, we interviewed students and teachers from Class C1. From these interviews, we identified two recurring themes.

\paragraph{\textbf{Theme 1. MAIC-UI expanded opportunities for active participation.}}

Participants perceived that MAIC-UI created more entry points for students to explore. One student noted becoming ``more willing to ask questions,'' whereas previously they had been reluctant due to concerns about how others might see them. Teachers observed that previously quiet students had ``completely opened themselves up in the smart online learning environment.''

\paragraph{\textbf{Theme 2. Visual interaction supported accessible understanding.}}

Both students and teachers reported that interactive visualization helped students understand processes and relationships. One teacher noted that students could more intuitively observe parameter changes ``through dragging and interaction'' when explaining force analysis and projectile motion. One student remarked that while they had previously only memorized that ``gravitational acceleration is 9.8,'' after interacting with the system they could observe that ``the higher the height, the longer the falling time,'' so that they ``understood it immediately.''

\section{Discussion}
\label{sec:discussion}

Our findings demonstrate that MAIC-UI successfully addresses key challenges in teacher-facing generative UI systems for educational content creation. We discuss key implications here; see Appendix~\ref{app:discussion} for the complete analysis.

\textbf{Balancing Automation and Pedagogical Control.}
Teachers want AI assistance but cannot compromise on pedagogical accuracy. MAIC-UI's two-stage pipeline separates content validation from visual polish, ensuring substance is verified before aesthetics are applied. This design reveals a broader principle: effective educational GenUI requires transparent communication of AI to help teachers leverage strengths while compensating for weaknesses.

\textbf{The Importance of Rapid Iteration Cycles.}
Iteration speed fundamentally shapes creative engagement. Our sub-10-second iteration cycles---enabled by Unified Diff-based incremental generation---transform authoring from a batch-oriented process into an interactive conversation. Delays exceeding approximately 10 seconds create cognitive discontinuities that disrupt creative flow.


\textbf{Addressing Equity in STEM Learning.}
MAIC-UI's benefits were not limited to high-performing students. Lower-performing students in the pilot class demonstrated larger score gains than their counterparts in control classes, and variance of score gains was substantially lower. This suggests interactive courseware may help address persistent achievement gaps by making abstract concepts accessible through multiple entry points.

\textbf{Limitations.}
MAIC-UI currently focuses on single-page simulations, limiting applicability for extended narratives. The lab study used proxy instructors rather than practicing K-12 teachers, and the classroom deployment occurred in a single Chinese public high school, requiring further investigation for generalizability. 

\section{Conclusion}

This paper presented MAIC-UI, a zero-code authoring system that enables educators to create interactive STEM courseware without programming. The system introduces three technical contributions: structured knowledge analysis for pedagogical rigor, a generate-verify-optimize pipeline, and Click-to-Locate editing with Unified Diff-based incremental generation achieving sub-10-second iteration cycles. Our three-month classroom deployment with 53 high school students demonstrated that MAIC-UI reduces editing iterations, improves learnability for non-programmers, and fosters learning agency---the pilot class achieved 9.21-point STEM gains compared to -2.32 points in control classes. Our findings suggest that well-designed interactive courseware can help address persistent achievement gaps in STEM education. Future work will explore template-based courseware support, additional disciplines, and long-term integration across diverse educational contexts.
\bibliographystyle{ACM-Reference-Format}
\bibliography{sample-base}
\appendix

\section{Detailed Discussion}
\label{app:discussion}

This appendix presents the complete discussion of implications and limitations that were summarized in Section~\ref{sec:discussion}.

\subsection{Implications for Teacher-Facing GenUI Systems}

\subsubsection{Balancing Automation and Pedagogical Control}

Our formative study revealed a fundamental tension in educational GenUI systems: teachers want AI assistance to reduce workload, but they cannot compromise on pedagogical accuracy. This finding aligns with prior work highlighting educators' concerns about AI-generated content misrepresenting scientific concepts~\cite{alasadi2023generative}. MAIC-UI addresses this tension through its two-stage generation pipeline, which separates content-aligned simulation creation from visual polish. This design ensures that pedagogical substance is validated before aesthetic enhancements are applied.

However, our findings also suggest that effective human-AI collaboration in educational contexts requires more than workflow design---it demands clear communication of AI capabilities and limitations. Participants in our study who understood that the system performed structured content analysis were better able to leverage its strengths while compensating for its weaknesses. Future systems should consider transparent visualization of the extraction and analysis process, helping teachers understand what content has been identified and how it will drive generation.

\subsubsection{The Importance of Rapid Iteration Cycles}

A key insight from our evaluation is that iteration speed fundamentally shapes creative engagement. When modifications require 200--600 seconds, as in baseline systems, teachers report losing their ``train of thought'' and abandoning refinement attempts. MAIC-UI's sub-10-second iteration cycles, enabled by Unified Diff-based incremental generation, transform the authoring experience from a batch-oriented process into an interactive conversation.

This finding has broader implications for GenUI design beyond education. Any domain requiring iterative refinement---such as data visualization, dashboard design, or document formatting---may benefit from incremental update mechanisms that preserve context while minimizing latency. The key principle is that creative flow requires feedback loops tighter than human working memory can comfortably sustain; delays exceeding approximately 10 seconds create noticeable discontinuities in cognitive engagement~\cite{doherty2018engagement}.

\subsubsection{Zero-Code Editing as End-User Programming}

MAIC-UI's Click-to-Locate interface represents a novel approach to end-user programming for web content. By allowing teachers to select elements visually and describe changes in natural language, the system bridges the gap between the rendered interface and its underlying code representation. This approach contrasts with traditional developer tools that require understanding of DOM structures and CSS selectors.

Our findings suggest that this design successfully embodies the EUP principle of enabling users to focus on \textit{what} they want to achieve rather than \textit{how} to implement it. Teachers could make precise modifications without writing code, yet the system preserved their ability to articulate specific intentions through natural language. Future GenUI systems for non-expert users should consider similar mechanisms for bridging visual and code representations, particularly in domains where users have strong domain expertise but limited programming knowledge.

\subsection{Implications for Procedural Knowledge Learning in STEM Education}

\subsubsection{Externalizing Procedural Thinking}

Traditional STEM instruction often relies on symbolic representations---equations, formulas, and static diagrams---that require students to mentally simulate dynamic processes. MAIC-UI's interactive simulations externalize these mental simulations, making procedural knowledge visible and manipulable. Our classroom deployment findings suggest this externalization is particularly beneficial for lower-performing students, who showed larger score gains and more even distribution of learning outcomes.

This pattern aligns with theories of distributed cognition, which posit that cognitive processes can be offloaded onto environmental structures~\cite{hollan2000distributed}. By making abstract relationships concrete and interactive, MAIC-UI reduces the working memory load required for understanding procedural concepts. Future educational technology design should consider how interactive representations can serve as ``cognitive scaffolds'' that make invisible thought processes visible and explorable.

\subsubsection{Fostering Learning Agency Through Interactive Exploration}

Our in-class deployment revealed that interactive courseware can shift classroom dynamics from teacher-centered presentation to student-centered exploration. Students reported becoming ``more willing to ask questions'' and engage actively with content, while teachers described ``playing with knowledge points together with students'' rather than ``explaining concepts to the blackboard.''

This transformation represents a shift from passive reception to active construction of knowledge, consistent with constructivist learning theories~\cite{bada2015constructivism}. However, our findings also highlight that such shifts depend on the quality of interactive design---poorly designed interactions may distract from learning rather than enhance it. The key appears to be alignment between the interactive affordances and the underlying procedural structure: when manipulations directly correspond to conceptual variables, exploration supports understanding; when they do not, interaction may become mere entertainment.

\subsubsection{Addressing Equity in STEM Learning}

Perhaps most significantly, our classroom deployment showed that MAIC-UI's benefits were not limited to already high-performing students. Lower-performing students in the pilot class demonstrated larger score gains than their counterparts in control classes, and the variance of score gains was substantially lower, suggesting more even distribution of learning benefits.

This finding suggests that interactive courseware may help address persistent achievement gaps in STEM education. By externalizing procedural knowledge and providing multiple entry points for engagement, such systems may make abstract concepts accessible to students who struggle with traditional symbolic representations. Future research should explore how such tools can be designed to maximize equity benefits, potentially through adaptive scaffolding or personalized interaction pathways.

\subsection{Limitations}

While our evaluation demonstrates MAIC-UI's effectiveness, several limitations should be acknowledged.

\subsubsection{System Limitations}

MAIC-UI currently focuses on generating single-page interactive simulations rather than comprehensive multi-page courseware. While this scope aligns with our goal of supporting focused procedural knowledge visualization, it limits applicability for topics requiring extended narrative or sequential lesson structures. Future work should explore how the Click-to-Locate editing paradigm can scale to multi-page experiences while maintaining rapid iteration cycles.

Additionally, the current implementation relies on vision-language models for content extraction, which may occasionally misinterpret specialized notation or complex diagrams. While our structured analysis prompt reduces these errors, domain-specific fine-tuning or hybrid approaches combining OCR with visual understanding may be necessary for optimal performance across diverse STEM disciplines.

\subsubsection{Evaluation Limitations}

Our lab user study employed proxy instructors (graduate students with teaching experience) rather than practicing K-12 teachers. While this approach enabled controlled comparison, the participant population may differ from our target users in terms of technical comfort, pedagogical training, and authentic classroom pressures. The three-month classroom deployment with actual high school teachers and students helps address this limitation, but longer-term studies examining sustained use and integration into regular teaching practice are needed.

Furthermore, our classroom deployment was conducted in a single school with a specific student population (physics-chemistry-biology track in a Chinese public high school). The generalizability of our findings to other educational contexts---including different countries, subject areas, or student demographics---requires further investigation. Cultural differences in teaching practices and student expectations may influence how interactive courseware is received and integrated.

\subsubsection{Technical Constraints}

The Unified Diff-based incremental generation depends on stable code structure to apply patches correctly. While our fuzzy context matching handles minor drift, substantial structural changes from extensive editing sessions may occasionally require full regeneration. Additionally, the current system requires internet connectivity for AI model access, limiting use in settings with poor connectivity or strict data privacy requirements. Future iterations could explore local model deployment or hybrid approaches that balance capability with accessibility.

\section{Formative Study Methodology}
\label{app:formative-methodology}

\subsection{Participant Recruitment}
\label{app:recruitment}

We recruited 6 participants for the formative study through purposive sampling from local university networks. The recruitment criteria included: (1) senior undergraduate or graduate students from top-tier Chinese universities, (2) at least one formal teaching experience (tutoring, teaching assistantship, or formal instruction), and (3) no prior exposure to MAIC-UI or similar AI-powered courseware generation systems.

\textbf{Recruitment channels:} Participants were recruited through departmental mailing lists, teaching center announcements, and snowball sampling from existing contacts. Potential participants completed a screening survey assessing their teaching experience, disciplinary background, and familiarity with educational technology tools.

\textbf{Participant demographics:}
\begin{itemize}
    \item P1: Graduate student, Computer Science, 2 years tutoring experience
    \item P2: Graduate student, Physics, 1 year teaching assistantship
    \item P3: Senior undergraduate, Mathematics, 3 years tutoring experience
    \item P4: Graduate student, Chemistry, 1.5 years teaching assistantship
    \item P5: Graduate student, Biology, 2 years tutoring experience
    \item P6: Senior undergraduate, Engineering, 2 years tutoring experience
\end{itemize}

All participants provided written informed consent and received a compensation of 200 RMB (approximately \$28 USD) for their participation.

\subsection{Interview Protocol}
\label{app:interview-protocol}

Each participant completed a 1-hour hands-on session with MAIC-UI's initial prototype, followed by a semi-structured interview lasting approximately 45 minutes. The interview protocol covered five main areas:

\textbf{1. Creation Experience (10 minutes)}
\begin{itemize}
    \item Walk me through your process of creating the interactive courseware.
    \item What aspects of the system felt intuitive or confusing?
    \item How did you decide what changes to make during the editing process?
\end{itemize}

\textbf{2. Perceived Learning Costs (5 minutes)}
\begin{itemize}
    \item How long do you think it would take to become proficient with this system?
    \item What background knowledge would teachers need to use this effectively?
    \item How does this compare to learning traditional courseware creation tools?
\end{itemize}

\textbf{3. Creative Amplification (10 minutes)}
\begin{itemize}
    \item How did the system support or constrain your creative ideas?
    \item Were there design ideas you wanted to implement but couldn't?
    \item How did the generated output compare to your initial vision?
\end{itemize}

\textbf{4. Classroom Integration (10 minutes)}
\begin{itemize}
    \item How do you envision using this tool in your actual teaching?
    \item What concerns would you have about using AI-generated materials in class?
    \item How might students respond to interactive versus traditional materials?
\end{itemize}

\textbf{5. Procedural Knowledge Visualization (10 minutes)}
\begin{itemize}
    \item How well did the system handle step-by-step procedures or processes?
    \item What types of content do you think would benefit most from interactive visualization?
    \item Can you describe a specific concept that would be difficult to teach without interactive elements?
\end{itemize}

Interviews were audio-recorded with participant consent and transcribed verbatim for thematic analysis.

\subsection{Data Analysis}
\label{app:data-analysis}

Two researchers independently coded the interview transcripts using deductive and inductive thematic analysis. We first developed a preliminary coding scheme based on our research questions and then refined it through iterative coding. Inter-rater reliability was calculated on 20\% of the transcripts, achieving Cohen's kappa of 0.82. Discrepancies were resolved through discussion until consensus was reached.

\section{Extended Participant Quotes}
\label{app:extended-quotes}

This section presents extended verbatim quotes from the formative study that were abbreviated or omitted from the main text due to space constraints.

\subsection{Knowledge Accuracy Concerns}

\textbf{P3 on pedagogical scientificity:}
\begin{quote}
    ``The main issue lies in how it represents knowledge. When you're teaching, you can't just present facts; you need to show the logic, the derivation, the scientific rigor. The system generates content quickly, but it doesn't always get the scientific concepts quite right. For example, when I asked it to explain Newton's laws, it gave a correct statement but missed the nuances that students typically struggle with. As a teacher, I need to verify every piece of generated content, which takes time.''
\end{quote}

\textbf{P4 on content verification needs:}
\begin{quote}
    ``Sometimes the knowledge it produces is simply incorrect. Not often, but even once is too much when you're teaching. Imagine standing in front of thirty students and presenting something wrong---your credibility is gone. So I would need to check everything carefully, maybe even rewrite portions. The speed is helpful, but only if the quality is there.''
\end{quote}

\textbf{P6 on content alignment:}
\begin{quote}
    ``The generated website didn't include the content I had specified. I gave it a PDF with specific examples I wanted to use, but the output focused on generic explanations instead. It felt like the system was making assumptions about what was important rather than actually using my materials. If I'm going to use this, it needs to respect my input---I'm the teacher, I know what my students need.''
\end{quote}

\subsection{Editing Experience}

\textbf{P3 on the difficulty of editing:}
\begin{quote}
    ``Modification and editing aren't that easy. You describe what you want changed, and sometimes AI pretends to understand but gets it wrong. Then you have to explain again, and again. It becomes this back-and-forth where you're not sure if the problem is your instructions or the system's understanding. After three or four tries, I started wondering if it would be faster to just do it myself.''
\end{quote}

\textbf{P5 on iteration requirements:}
\begin{quote}
    ``In cases like this, you might need to edit three to four times. The first generation gives you something close but not quite right. The second edit gets closer. By the third or fourth iteration, it's usually usable. But that's a lot of waiting---each regeneration takes time, and you're sitting there hoping this time it will be right. For simple text changes, it's frustrating.''
\end{quote}

\textbf{P6 on granular control limitations:}
\begin{quote}
    ``I kept trying to change the website---specifically, I wanted to adjust the layout of a particular section to better match my teaching flow. But in the end, it didn't follow my instructions. It changed something, but not what I asked for. When it comes to detailed issues, like breaking a procedure into several steps with specific visual hierarchy, it may not handle those modifications well. I felt like I had to accept whatever it gave me.''
\end{quote}

\subsection{Passive vs. Active Learning}

\textbf{P4 on traditional teaching limitations:}
\begin{quote}
    ``Traditional PPT and blackboard teaching---it's fixed content. You write it, you present it, and students watch. They may find it boring and fail to concentrate, especially for abstract concepts. There's no exploration, no discovery. With interactive tools, you can generate random parameters, let students experiment. They become more concentrated because each interaction is unique. They get randomized experiences instead of the same example every time.''
\end{quote}

\textbf{P2 on active learning benefits:}
\begin{quote}
    ``Traditional teaching requires feeding knowledge to students' mouths before they attempt it. You're essentially saying, `Trust me, this is how it works.' But interactive tools offer students a buffet---they can explore in an immediately actionable way. Everyone is lazy, but if you give them a better opportunity to try, they will be more active. When students manipulate variables themselves and see the results, they own that understanding. It's not just received knowledge anymore.''
\end{quote}

\subsection{Theory-Practice Gap}

\textbf{P4 on the knowledge-application disconnect:}
\begin{quote}
    ``What they learn from textbooks and real-world scenarios have a gap. Textbooks present idealized situations---frictionless surfaces, perfect gases, ideal circuits. But real-world scenarios are messy. If students don't think actively about how to bridge this gap themselves, they won't make the connection. The knowledge they learn is `dead'---it's memorized but not truly understood. But when facing more flexible applications in exams or real life, they become confused because they've never seen the principles in action.''
\end{quote}

\textbf{P2 on the challenge of abstraction:}
\begin{quote}
    ``Problems are written elaborately, with complex scenarios and multiple steps, but the underlying knowledge is simple---some students cannot cross this chasm. They get lost in the problem description and can't see the basic principle underneath. Visualization helps bridge that gap. If they can see the force vectors, see the motion, suddenly the abstract symbols make sense. Without that connection, they're just memorizing problem-solving templates.''
\end{quote}

\textbf{P5 on intuitive understanding:}
\begin{quote}
    ``Students lack intuitive feelings for concepts when first encountering them. They can recite definitions, but they don't `feel' what the concept means. For velocity, they can say $v = d/t$, but do they intuitively understand what changing velocity feels like? Interactive visualizations give them that feeling---they can see it, manipulate it, experience it. That builds the intuition that textbooks alone can't provide.''
\end{quote}

\section{Full Questionnaire Items}
\label{app:questionnaire}

This section presents the complete questionnaire items used in the lab user study (Section~\ref{sec:lab_user_study}).

\subsection{Post-Task Questionnaire (RQ2: Usability and Editing Experience)}

Participants rated the following items on a 5-point Likert scale (1 = Strongly Disagree, 5 = Strongly Agree):

\textbf{Learnability and Usability Items}
\begin{enumerate}
    \item[I1.] \textbf{Learnability:} I could quickly learn to use the system effectively.

    \item[I2.] \textbf{Time Cost:} The time required to create and refine the courseware was acceptable.

    \item[I3.] \textbf{Editing Controllability:} I could precisely control what changes were made to the generated courseware.

    \item[I4.] \textbf{Usage Preference:} I would prefer using this system over traditional courseware creation methods.
\end{enumerate}

\textbf{Perceived Quality Items (MAIC-UI condition only)}
\begin{enumerate}
    \item[Q1.] \textbf{Layout Intuitiveness:} The layout of the generated webpage is intuitive and easy to follow.

    \item[Q2.] \textbf{Attention Attraction:} The visual design attracts and maintains learner attention.

    \item[Q3.] \textbf{Concept Accuracy:} The webpage accurately presents key concepts and procedural steps.

    \item[Q4.] \textbf{Content Coverage:} The webpage covers all key teaching points from the source material.

    \item[Q5.] \textbf{Language Clarity:} The instructional language used is clear and appropriate for learners.

    \item[Q6.] \textbf{Concept Intuitiveness:} The presentation of key concepts is intuitive and easy to understand.
\end{enumerate}

\subsection{Demographic Questionnaire}

\begin{enumerate}
    \item What is your current academic status? (Undergraduate / Master's student / Doctoral student / Other)
    \item What is your disciplinary background? (Social Sciences / Computer Science / Basic Sciences / Engineering / Other)
    \item How many hours of teaching experience do you have? (tutoring, TAship, or formal instruction)
    \item How would you rate your programming experience? (None / Beginner / Intermediate / Advanced)
    \item How would you rate your experience with AI tools (e.g., ChatGPT)? (None / Beginner / Intermediate / Advanced)
    \item Have you created educational courseware before? (Yes / No)
\end{enumerate}

\subsection{Post-Study Interview Guide (Lab Study)}

\textbf{Usability and Experience:}
\begin{itemize}
    \item What aspects of the system were most helpful during courseware creation?
    \item What aspects were most frustrating or difficult?
    \item How did you decide what changes to request during the editing phase?
    \item How did the editing experience compare to your expectations?
\end{itemize}

\textbf{Perceived Quality:}
\begin{itemize}
    \item How satisfied were you with the final courseware quality?
    \item What additional features or capabilities would you want?
    \item How would you use this tool in your actual teaching context?
\end{itemize}

\section{Detailed Implementation}
\label{app:implementation}

This section provides technical implementation details of the MAIC-UI system.

\subsection{System Architecture}

MAIC-UI follows a client-server architecture with the following components:

\textbf{Frontend:}
\begin{itemize}
    \item \textbf{Framework:} React 18 with TypeScript 5.0
    \item \textbf{State Management:} Zustand for global state, React Query for server state
    \item \textbf{UI Components:} Custom components with Tailwind CSS for styling
    \item \textbf{Code Editor:} Monaco Editor for HTML/CSS viewing
    \item \textbf{Preview:} Sandboxed iframe with bi-directional message passing
\end{itemize}

\textbf{Backend:}
\begin{itemize}
    \item \textbf{Framework:} Python 3.11 with FastAPI
    \item \textbf{Database:} SQLite with SQLAlchemy ORM
    \item \textbf{Task Queue:} Celery with Redis for async processing
    \item \textbf{File Storage:} Local filesystem with CDN support
\end{itemize}

\subsection{AI Model Configuration}

\textbf{Multi-modal Analysis:}
\begin{itemize}
    \item \textbf{Model:} GLM-4.6V (Zhipu AI)
    \item \textbf{Parameters:} temperature=0.2, max\_tokens=4096
    \item \textbf{Fallback:} GLM-4.5V
\end{itemize}

\textbf{Text Generation:}
\begin{itemize}
    \item \textbf{Model:} GLM-4.7 (Zhipu AI)
    \item \textbf{Parameters:} temperature=0.3, max\_tokens=8192
    \item \textbf{Fallback:} GLM-4.6
\end{itemize}

\subsection{Structured Analysis Prompt}

The multi-modal analysis uses the following structured prompt template:

\begin{lstlisting}[language=python]
ANALYSIS_PROMPT = """
Analyze the provided educational document and extract the following
information in JSON format:

1. Main Topics: List 3-5 broad subject areas covered
2. Key Concepts: Specific terminology and principles students must master
3. Learning Objectives: Measurable outcomes students should achieve
4. Prerequisite Knowledge: Foundational concepts required beforehand
5. Procedural Concepts: Step-by-step processes suitable for simulation
   - Name of the procedure
   - List of steps
   - Adjustable parameters
6. Subject Area: One of [Physics, Chemistry, Biology, Math, Geography, Other]
7. Grade Level: One of [Primary, Middle, High, Undergraduate, Graduate]

Focus on identifying content that would benefit from interactive
visualization. Be precise and comprehensive.

Response format: Valid JSON only, no markdown formatting.
"""
\end{lstlisting}

\subsection{Two-Stage Generation Pipeline}

\textbf{Stage 1 Prompt (Content-Aligned Simulation):}
\begin{lstlisting}[language=python]
STAGE1_PROMPT = """
Generate an interactive HTML/JavaScript simulation based on the
following educational content:

Subject: {subject_area}
Key Concepts: {key_concepts}
Procedural Concepts: {procedural_concepts}

Requirements:
1. Left panel: Step-by-step process display with current step highlighting
2. Right panel: Interactive controls for adjusting parameters
3. Real-time coupling between process and simulation panels
4. Scientific accuracy is paramount---verify all formulas and relationships
5. Include explanatory tooltips for technical terms
6. Use vanilla JavaScript (no external dependencies)
7. Responsive layout for tablet devices (min-width: 768px)

Generate complete, valid HTML with embedded CSS and JavaScript.
"""
\end{lstlisting}

\textbf{Stage 2 Prompt (Visual Polish):}
\begin{lstlisting}[language=python]
STAGE2_PROMPT = """
Apply visual polish to the following HTML simulation:

Current HTML: {stage1_html}
Theme: {theme_config}

Enhancements to apply:
1. Apply theme colors consistently (primary: {primary}, accent: {accent})
2. Improve typography hierarchy
3. Add smooth animations for state transitions
4. Ensure consistent spacing and alignment
5. Validate all HTML structure
6. Maintain all interactive functionality

Return complete polished HTML.
"""
\end{lstlisting}

\subsection{Click-to-Locate Implementation}

The Click-to-Locate feature is implemented through DOM-aware element tracking:

\begin{lstlisting}[language=python]
// Frontend element selection
function handleElementClick(event: MouseEvent) {
  const element = event.target as HTMLElement;
  const selector = {
    xpath: getXPath(element),
    cssSelector: getCSSSelector(element),
    elementHtml: element.outerHTML.substring(0, 500),
    boundingBox: element.getBoundingClientRect()
  };

  // Send to backend with instruction
  const editRequest = {
    element_selector: selector,
    instruction: userInstruction,
    context_html: document.documentElement.outerHTML
  };

  streamEditRequest(editRequest);
}

// Generate XPath for element
def getXPath(element: HTMLElement): string {
  if (element.id) return `//*[@id="${element.id}"]`;
  const path = [];
  while (element.parentElement) {
    const siblings = Array.from(element.parentElement.children)
      .filter(e => e.tagName === element.tagName);
    const index = siblings.indexOf(element) + 1;
    path.unshift(`${element.tagName}[${index}]`);
    element = element.parentElement;
  }
  return `/${path.join('/')}`;
}
\end{lstlisting}

\subsection{Unified Diff Processing}

The system uses Unified Diff format for incremental updates:

\begin{lstlisting}[language=python]
import difflib

def generate_unified_diff(original: str, modified: str) -> str:
    """Generate unified diff between original and modified HTML."""
    original_lines = original.splitlines(keepends=True)
    modified_lines = modified.splitlines(keepends=True)

    diff = difflib.unified_diff(
        original_lines,
        modified_lines,
        fromfile='original.html',
        tofile='modified.html'
    )

    return ''.join(diff)

def apply_diff(original: str, diff: str) -> str:
    """Apply unified diff to original HTML."""
    # Fuzzy matching for context drift
    lines = original.splitlines()
    diff_lines = diff.splitlines()

    result = []
    i = 0
    for line in diff_lines:
        if line.startswith('---') or line.startswith('+++'):
            continue
        elif line.startswith('@@'):
            # Parse hunk header
            continue
        elif line.startswith('-'):
            # Remove line (with fuzzy matching)
            continue
        elif line.startswith('+'):
            # Add line
            result.append(line[1:])
        else:
            # Context line
            result.append(line)

    return '\n'.join(result)
\end{lstlisting}

\subsection{Performance Optimization}

\textbf{Response Time Targets:}
\begin{itemize}
    \item Initial generation: 30-60 seconds
    \item Edit iterations: $<$10 seconds (p50: 6.2s, p95: 8.8s)
    \item Element selection: $<$100ms
    \item Preview rendering: $<$500ms
\end{itemize}

\textbf{Optimization Strategies:}
\begin{enumerate}
    \item \textbf{Token Reduction:} Unified Diff format reduces output tokens by $\sim$90\% vs. full HTML regeneration
    \item \textbf{Streaming:} Progressive rendering with Server-Sent Events (SSE)
    \item \textbf{Caching:} Analysis results cached for 24 hours, generation templates cached indefinitely
    \item \textbf{Async Processing:} Non-blocking I/O for concurrent requests
    \item \textbf{Connection Pooling:} Keep-alive connections to AI API endpoints
\end{enumerate}

\subsection{Error Handling and Fallbacks}

The system implements a graceful degradation strategy:

\begin{enumerate}
    \item \textbf{Analysis Failure:} Return basic metadata extraction; allow manual input
    \item \textbf{Stage 1 Failure:} Fall back to single-pass generation with reduced validation
    \item \textbf{Stage 2 Failure:} Apply basic CSS styling to Stage 1 output
    \item \textbf{Both Stages Fail:} Return emergency template with user-friendly error message
    \item \textbf{Edit Failure:} Retry with expanded context; fall back to full regeneration after 3 failures
\end{enumerate}

\end{document}